# *asanAI*: In-Browser, No-Code, Offline-First Machine Learning Toolkit

**Norman Koch**[1,2], **Siavash Ghiasvand**[1,2]

[1]Center for Interdisciplinary Digital Sciences (CIDS), Technische Universität Dresden, Germany
[2]Center for Scalable Data Analytics and Artificial Intelligence (ScaDS.AI) Dresden/Leipzig. Germany
norman.koch@tu-dresden.de, siavash.ghiasvand@tu-dresden.de

## Abstract

Machine learning (ML) has become crucial in modern life, with growing interest from researchers and the public. Despite its potential, a significant entry barrier prevents widespread adoption, making it challenging for non-experts to understand and implement ML techniques. The increasing desire to leverage ML is counterbalanced by its technical complexity, creating a gap between potential and practical application. This work introduces *asanAI*, an offline-first, open-source, no-code machine learning toolkit designed for users of all skill levels. It allows individuals to design, debug, train, and test ML models directly in a web browser, eliminating the need for software installations and coding. The toolkit runs on any device with a modern web browser, including smartphones, and ensures user privacy through local computations while utilizing WebGL for enhanced GPU performance. Users can quickly experiment with neural networks and train custom models using various data sources, supported by intuitive visualizations of network structures and data flows. *asanAI* simplifies the teaching of ML concepts in educational settings and is released under an open-source MIT license, encouraging modifications. It also supports exporting models in industry-ready formats, empowering a diverse range of users to effectively learn and apply machine learning in their projects. The proposed toolkit is successfully utilized by researchers of ScaDS.AI to swiftly draft and test machine learning ideas, by trainers to effectively educate enthusiasts, and by teachers to introduce contemporary ML topics in classrooms with minimal effort and high clarity.

## Introduction

Machine learning is now a fundamental part of our daily lives, powering the devices we use and enhancing the services we rely on. As it transforms our world, many people want to understand its basic principles, potential, and limitations. Given its widespread presence, having some basic knowledge of machine learning is increasingly beneficial. While courses on machine learning and artificial intelligence are now part of many school curriculums, the rapid advancements in AI outpace traditional educational tools. Consequently, there is an urgent need to develop educational toolkits to support programs that teach ML concepts to the next generation. To address this challenge and pave the way for democratizing Machine Learning, *asanAI*[1], a no-code, browser-based, offline-first, and user-friendly ML toolkit is introduced. *asanAI* enables users without prior programming knowledge to design, debug, train, test, and understand machine learning models.

*asanAI* is accessible on any device with a JavaScript-enabled modern browser, including tablets and smartphones. While existing machine learning toolkits often have high entry barriers and requirements such as familiarity with programming languages, complex installations, or reliable internet connection, *asanAI* focuses on removing these obstacles and allows a diverse spectrum of users to explore machine learning models without prior requirements simply via a web address[2]. All settings, configurations, and design elements are accessible through touch gestures or a mouse pointer, making the proposed toolkit particularly suitable for beginners who are more accustomed to touchscreen interfaces, such as those found on mobile devices.

*asanAI* supports various import functions and allows trained models to be exported into widely accepted formats to ensure compatibility and interoperability. It places significant emphasis on user privacy, which is achieved through local computation on the user's machine. Once the *asanAI* web page is fully loaded, the internet connection is no longer required, and its full features[3] can be safely used offline.

The proposed toolkit allows users to experiment with neural networks and design them through trial and error. It offers a variety of visualizations to help users understand the internal structure of networks better. For simple networks, the underlying formulas and calculations are dynamically displayed. When possible, *asanAI* uses WebGL hardware acceleration to perform calculations on the local GPU for improved performance. For high-performance applications, the designed model and its data can be exported as a standalone model for execution in a Python environment. As an open-source software, users can modify and adapt *asanAI* to their needs, making it a valuable tool for both beginners and experts in the field.

This document continues by summarizing the relevant lit-erature and highlighting the necessity for the development of *asanAI*. Next, the main design concepts that underpin the proposed toolkit are introduced. The subsequent section discusses the functionality and usability aspects of *asanAI*, followed by explanation of the key factors considered during the implementation of the toolkit. Finally we conclude the proposed work and outline potential future directions.

---

[1]Derived from Persian "asan" meaning easy or simple, and "AI" for Artificial Intelligence.
[2]*asanAI* toolkit: https://asanai.scads.ai
[3]With the exception of loading example data.

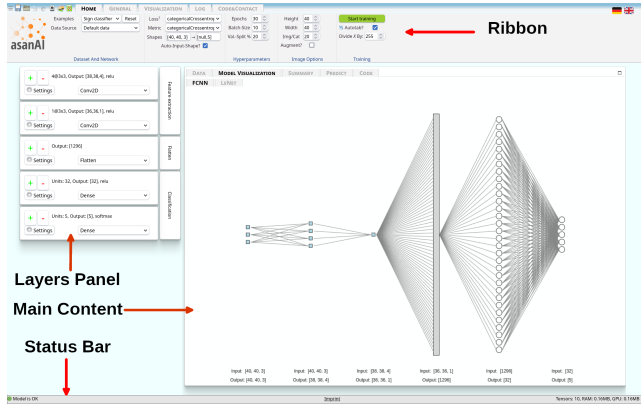

Figure 1: Overview of *asanAI* graphical user interface

## Related work

The development of no-code, browser-based machine learning toolkits such as *Teachable Machine* (Carney et al. 2020), *Lobe* (Microsoft 2021), *ml5.js* (Shiffman 2020), and *TensorFlow.js* (Smilkov et al. 2019) has accelerated in recent years to make machine learning more accessible to non-experts, aiming to enable faster experimentation and learning of machine learning concepts without requiring programming skills. Both *Teachable Machine* and *Lobe*, developed by Google and Microsoft respectively, allow for the training of image, audio, and other classification models. Although their usage is straightforward, users are limited to the existing use cases and models available in these tools. Furthermore, toolkits such as *Teachable Machine* require an active and reliable Internet connection, as all computations are performed on the server side, which also raises concerns regarding their usage with sensitive data, particularly in schools. Other tools, such as *Lobe*, are only executable on specific operating systems (e.g., Microsoft Windows), while *ml5.js* and *TensorFlow.js* are JavaScript libraries for creating and training neural networks in the browser, although using these libraries necessitates adequate programming skills.

Browser-based machine learning toolkits offer significant advantages in accelerating the learning process of ML concepts. These platforms can provide immediate visual feedback on model performance, facilitate rapid experimentation with various architectures and hyperparameters, eliminate complex setup procedures, and allow users to grasp core ML concepts before delving into programming intricacies (Lane 2024). However, these tools are not without limitations. Technical constraints restrict the complexity of models and the size of datasets that can be processed. Additionally, they often lack the fine-grained control offered by traditional programming approaches. Perhaps most critically, the transition from these simplified environments to production-grade ML workflows can be challenging, if not impractical in some cases. The contrast between accessibility and practicality creates both opportunities and challenges in the rapidly changing field of machine learning education and application development.

## Design Concepts

The primary objective of this work is to develop a machine learning toolkit that is accessible to a broad audience, aiming to (1) educate users on neural network principles, (2) lower entry barriers for novices, and (3) simplify the design, training, and testing processes of various machine learning models for advanced users. Another key objective of this work was prioritizing user privacy by design and to facilitate a seamless transition of developed models from conceptual stages to real-world applications and environments.

In response to these objectives, *asanAI* was conceived as an offline-first toolkit, operable on virtually any end-user device. It features a no-code, browser-based graphical user interface, eliminating the need for prior programming expertise. This design choice ensures that any device with a modern internet browser can access and utilize the toolkit.

*asanAI* places a strong emphasis on model explainability and interpretability, crucial aspects in modern ML applications, via providing various intuitive forms of data representation and visualizations. It also facilitates a smoother transition to programming-based ML frameworks via automatic generation of programming codes, bridging the gap between no-code environments and professional ML development.

This holistic approach establishes *asanAI* as a significant advancement in accessible yet powerful machine learning tools for education and prototyping, with a particular emphasis on interactivity, interoperability, and intuitiveness. Figure 1 provides an overview of the *asanAI* graphical user interface.

**Interactivity** Interactive software with immediate response provides a more engaging, personalized, and efficient experience compared to action button-based interfaces. This approach leads to higher engagement levels and better information retention among users. To deliver such an engaging experience, the *asanAI* user interface offers direct access to all configurable parameters and immediate feedback for every user interaction. Furthermore, the graphical user interface is designed to be easily operable using touch screen inputs, in addition to standard mouse and keyboard inputs. This ensures a smooth user experience on mobile devices like tablets and smartphones. It also makes interactions with asanAI simple and intuitive for any age group, regardless of their previous experience with computers.

**Interoperability** To ensure a high level of compatibility and enable seamless further usage of models, *asanAI* leverages the features of widely used software and protocols. At every stage of the neural network's design, training, and testing, *asanAI* enables users to export the entire model, including its architecture, weights, and data set. The exported data can be edited, shared, and seamlessly incorporated into further scenarios under a permissive license (CC-BY[4]), ensuring flexibility and interoperability throughout the model development life cycle.

Furthermore, at each stage of the design process, *asanAI* automatically generates and allows for the exportation of the equivalent Python code for the designed model. This functionality ensures model readiness and enables users to seamlessly execute their drafted machine learning model in any preferred native programming environment (e.g., a HPC system) without additional configuration.

Another key objective is to ensure *asanAI*'s smooth operation across all devices. During development, various measurement and optimization techniques were employed to identify and rectify performance bottlenecks. Extensive caching and memoization strategies were implemented to optimize model creation and compilation processes. Furthermore, the hardware accelerators available on the local device, such as GPUs, are automatically detected and utilized to enhance overall performance.

**Intuitiveness** The *asanAI* is structured hierarchically, mirroring the inherent nature of neural networks. This design adopts a default granularity at the layer level, allowing for individual definition and adjustment of each layer. The graphical user interface features a layer panel, displayed on the left side of Figure 1, which presents all the layers constituting the model along with their corresponding configurable parameters. By utilizing this interface, the necessary data structures are generated, and then underlying functions are called to instantiate the model.

Changes made in the graphical user interface result in an immediate update to the model. It is important to note that only one model can be loaded per page, as it is maintained as a singleton in browser memory. However, multiple tabs or browser windows can be opened, each containing a separate and independent model. Each layer, depending on its type, has a distinct set of associated options. For example, the `dense` layer type provides options such as whether it can be trained, the inclusion of bias, the number of units, the activation function, the initializers for both the kernel and bias, the regularizers for the kernel and bias, its visualization, and its data type. All these options are automatically selected within the GUI based on the chosen layer type.

Moreover, *asanAI* incorporates diverse instant verification methods to aid users in their design decisions and parameter selection. The level of this assistance can be calibrated by users to align with their expertise. To further assist novice users in leveraging their custom data and to reduce initial barriers to entry, *asanAI* automatically determines the model's shape. When this option is activated, depending on the input data type, *asanAI* can accurately detect the network's input shape. For instance, when dealing with images, the system will automatically present options for adjusting the widths and heights of images. For custom data types, such as CSV files, the input shape is determined by the number of input values. Additionally, in categorization tasks, the last layer of the network can automatically adjust to accommodate the number of categories corresponding to the uploaded images or the number of values present in the target set of the provided input data.

*asanAI* is provided as a multilingual tool. Users can select their preferred language from the options in the top right corner of the screen, which instantly translates the entire graphical user interface into the chosen language. User preferences can be saved in browser cookies, ensuring that the preference persists even after page reloads.

## Usability of *asanAI*

The proposed toolkit is designed to prioritize usability while providing users with full accessibility and flexibility to configure the model settings according to their preferences. As such, the software is designed with sane default settings, which ensure that users can easily navigate and perform the necessary functions without prior knowledge. Additionally, users can customize the software settings as desired. This design principle is aimed at improving the software's ease-of-use, thus providing a more efficient and streamlined user experience. In this way, *asanAI* is intended to be accessible to a wider audience, including those who may have limited to no experience or familiarity at all with machine learning tools.

**Layer Groups** In a neural network, a layer is defined as a group of nodes (neurons) that work collaboratively at a particular depth within the network. Each layer processes incoming data and transmits the resulting output to the subsequent layer. The typical architecture of a neural network includes three primary types of layers: the input layer, hidden layers, and the output layer. To enhance user understanding, asanAI categorizes these layers based on their functionality, such as "Feature Detection" or "Classification." Each layer has its own individual set of configuration parameters, which are intuitively accessible from the vertical layer panel on the left side of the web GUI as shown in figure 1.

**Operational Mode** Users at different skill levels have distinct needs when it comes to software usability. To cater to this diversity, *asanAI* provides three operational modes: Introductory, Beginner, and Expert. By default, *asanAI* launches in Beginner mode, which is suitable for users with basic machine learning knowledge. This mode strikes a balance between guidance and flexibility, assisting users in their choices while preventing them from making critical errors. It restricts the selection of invalid layer types and automatically corrects changes that could potentially break the configured model. This approach allows users to experiment and learn without the risk of creating non-functional models.

For those with advanced skills and requirements, the Expert mode offers a more sophisticated experience. It provides users with greater control over various aspects of model creation, including layer types, structures, input shapes, and other advanced features. While it offers less direct assistance compared to the Beginner mode, it allows for extensive experimentation and the ability to test various configuration combinations. This freedom comes with the possibility of creating models that may not function correctly however, it provides the flexibility that experienced users require.

---

[4]https://creativecommons.org/licenses/by/4.0/deed.de/

The Introductory mode streamlines the interface by hiding all configurations and instead provides a thorough explanation of machine learning, including a brief demonstration of an image classification model. Both the interactive introduction scenario and the classification model can be tailored to meet the audience's needs.

**Ribbon** In addition to the vertical 'Layer' panel in the *asanAI* graphical user interface, which provides access to individual layer settings, a horizontal Ribbon at the top of the screen offers access to various global settings for model design, training, data processing, visualization, and debugging. As illustrated in figure 2, the Ribbon is organized into multiple tabs, each offering specific group of configurations.

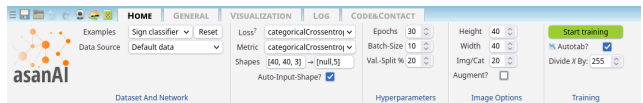

Figure 2: Ribbon; with the most commonly used options

The settings available on each tab vary depending on certain factors such as the input shape of the neural network and the operational mode of software. On the top left corner of the screen, the access to general functionalities are provided. The *Save/Open Model* options allow users to store their neural network configurations for future use or load pre-existing models. The *Undo* and *Redo* buttons enable users to navigate through their previous actions, facilitating experimentation with different configurations. Using the Camera button, training data can be streamed directly from a webcam, which is useful for tasks such as gesture recognition or object detection with custom objects. Finally, the Help button provides access to documentation, interactive tutorials, and other resources, ensuring users can find the information they need to make the most of the platform's features. When necessary, the Ribbon can be collapsed to maximize screen space.

**Live Training Progress** Once the training of a model begins, a series of live visualizations update continuously at the end of each batch or epoch, offering a real-time overview of the training progress. The Training tab presents various configurable visualizations, including loss and accuracy progress within each batch and epoch, time spent per batch, and a confusion matrix. The fully interactive live plots enable users to closely examine the training behavior, facilitating interactive debugging of the designed model.

**Immediate inference** At every stage after beginning the training of a model, the performance of model inference can be tested using the features available on the Predict tab, which automatically adapts itself based on the input data shape. For image-like inputs, users can test the model performance using local image files, drawn figures using a mouse pointer or touch input, or from a live webcam stream. For other custom inputs, users can enter a string representing their custom data structure, such as `[[true, false]]`.

**Visualization** To enhance the comprehensibility and explainability of the designed model, the *asanAI* GUI offers a range of visualization options. These visualizations enable users to gain a comprehensive understanding of the overall connectivity of neurons, explore each individual layer of the network, and monitor the training and prediction progress of models at their desired level of detail.

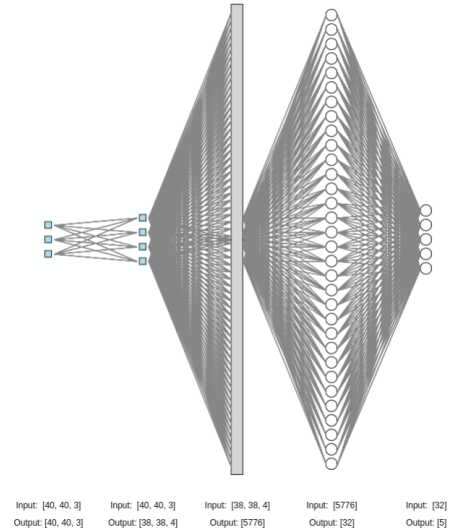

Figure 3: FCNN

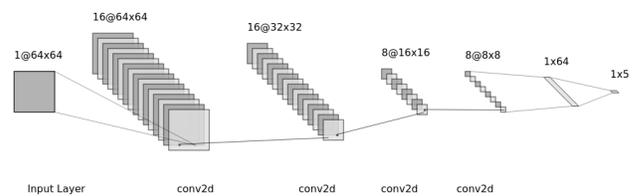

Figure 4: LeNet (LeNail 2019).

The FCNN[5] mode is depicted in figure 3, showcasing the network with neurons and filters in each layer, all fully connected. LeNet[6], illustrated in figure 4, presents the network as a sequence of filters, detailing their respective numbers and sizes. Additionally, intermediate layer visualizations, as shown in figure 5, allow for the examination of the input and output, as well as the kernels for convolutions, for every layer that possesses an image-like shape. Finally, loss visualizations provide insights into the performance of different loss functions when applied to custom data.

To complement the overarching visualizations that provide a comprehensive view of the model, several fine-granularity visualizations are provided, including Feature Map, GradCam, and Math-mode. For nearly all layer types, a "Visualize this layer" button is accessible in the layers settings, which calculates the input that most excites the output neurons of that layer. This functionality generates images that illustrate how the network interprets the input data to optimally activate specific neurons (Feature Map). Instead of training the network, this method involves fixing the neural network and training the input images to maximize the outputs (Zeiler and Fergus 2013). Figure 6 illustrates an example of such Feature Map.

---

[5] Reflects its fully connected architecture.

[6] Reflects the inspiration by original style from LeCun in 1998.

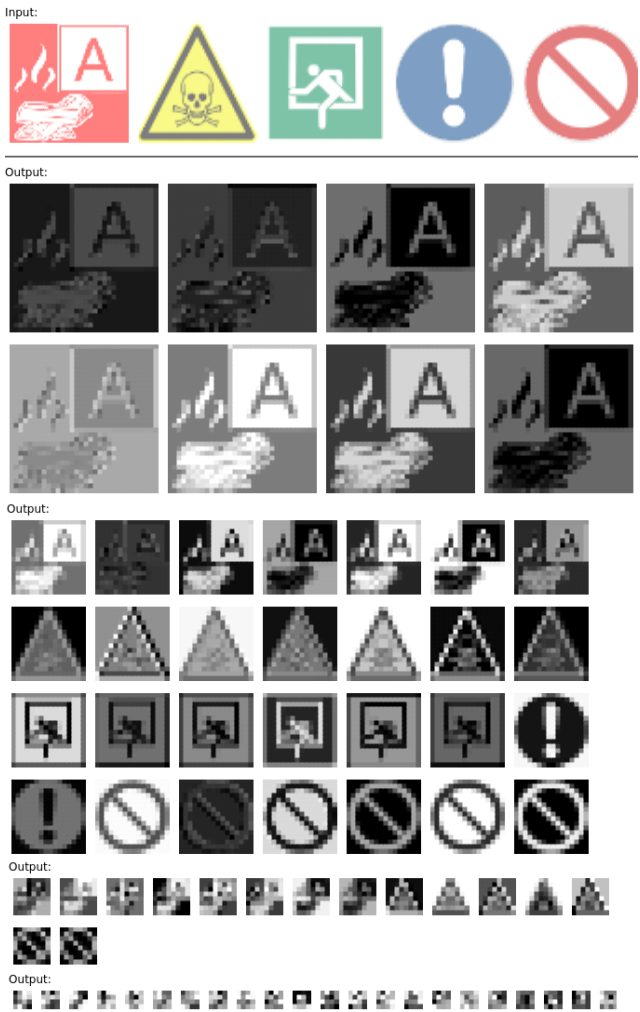

Figure 5: Inputs and outputs (data flow) of each layer

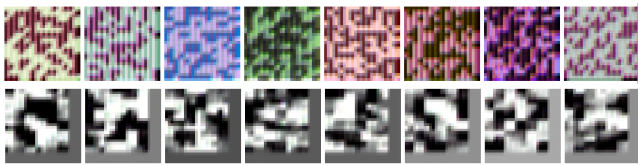

Figure 6: Feature Maps of the first- and last layer

GradCAM visualizations indicate which areas of an input image contribute most significantly to the classification of a specific output (Selvaraju et al. 2016). These visualizations resemble heatmaps, where the most yellow regions correspond to the areas that most activate the highest-scoring category, as illustrated in figure 7.

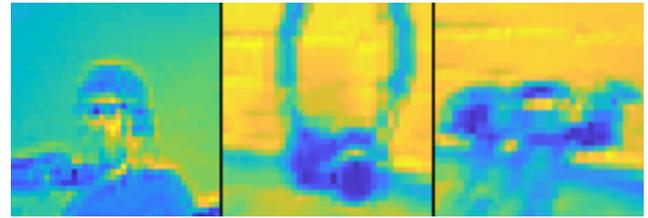

Figure 7: GradCAM visualization of a face, a headphone, and a car in real-time from live stream

The Math-mode feature of *asanAI* is designed mainly for simple neural networks where the input tensor is a vector and all layers are either Dense, Flatten, Reshape, Dropout, an activation layer, a Normalization layer, or a noise layer. The Math tab displays the precise equations computed by the underlying library, enabling users to follow and replicate these calculations manually, and achieve the same results. An example of the Math-mode visualization is shown in figure 8. In addition to educating users and enhancing the explainability of the model training process, the Math tab also provides a real-time visualization of the training progress. As the network trains, the visualizations are updated after each batch, color-coding the model weights: red for decreases, green for increases, and black for values that remain unchanged since the last batch.

**Activation functions:**

$$\mathrm{elu}(x) = \begin{cases} x & x \geq 0 \\ \alpha(e^x - 1) & x < 0 \end{cases}$$

$$\mathrm{sigmoid}(x) = \sigma(x) = \frac{1}{1+e^{-x}} \quad \text{(Lower-limit: 0, Upper-limit: 1)}$$

**Loss:**

$$\mathrm{MSE} = \frac{1}{n}\sum_{i=1}^{n}(y_i - \hat{y}_i)^2$$

**Layers:**

Layer 0 (dense): $h_{\text{Shape: }[8]} = \underbrace{\mathrm{elu}}_{\text{Activation}} \left( \underbrace{\begin{pmatrix} x_0 \\ x_1 \end{pmatrix}}_{\text{Input}} \times \underbrace{\begin{pmatrix} -1.02406 & 0.32688 & 0.34863 & 0.6548 & -0.78013 & 0.53626 & -0.10679 & -0.21833 \\ 0.51506 & 0.69122 & 0.79579 & 0.8052 & -0.39531 & 1.06088 & -0.24776 & -0.13844 \end{pmatrix}}_{\text{Kernel}^{2 \times 8}} + \underbrace{\begin{pmatrix} 0.32473 \\ -0.3725 \\ -0.41237 \\ -0.46124 \\ -0.34064 \\ -0.33214 \\ 0.35142 \\ 0.30676 \end{pmatrix}}_{\text{Bias}} \right)$

Layer 1 (dense): $h'_{\text{Shape: }[4]} = h_{\text{Shape: }[8]} \times \underbrace{\begin{pmatrix} -0.33738 & -1.04711 & 0.47883 & -0.82863 \\ 0.97371 & 0.53774 & 0.2409 & 0.56439 \\ 0.7242 & -0.07666 & -0.82074 & -0.03789 \\ 0.84993 & 0.21511 & -0.97678 & -0.36659 \\ 0.63139 & 1.06509 & 0.00914 & 0.74211 \\ 0.85483 & 1.07987 & -0.03204 & 0.50959 \\ -0.2892 & -1.14136 & 0.79062 & -0.75582 \\ -0.92379 & 0.06417 & 0.74758 & -0.28107 \end{pmatrix}}_{\text{Kernel}^{8 \times 4}} + \underbrace{\begin{pmatrix} -0.47762 \\ -0.40988 \\ 0.74818 \\ -0.57974 \end{pmatrix}}_{\text{Bias}}$

Layer 2 (dense): $\underbrace{(y_0)}_{\text{Output}} = \underbrace{\mathrm{sigmoid}}_{\text{Activation}} \left( h'_{\text{Shape: }[4]} \times \underbrace{\begin{pmatrix} 1.13237 \\ 0.53601 \\ -1.05261 \\ 1.36825 \end{pmatrix}}_{\text{Kernel}^{4 \times 1}} + \underbrace{(-0.22849)}_{\text{Bias}} \right)$

Figure 8: MathMode; live progress via color-coded values

**Own data import** *asanAI* currently offers three fundamental types of data importers: CSV, Image, and Tensors. The CSV importer processes a given CSV file using a specified separator. The first line is treated as the column title list. Users can select which column serves as the input and which as the target set. Additionally, the data in each column can be easily divided by any number (e.g., for normalization purposes). A live preview showcases the resulting tensors. The importer can automatically adjust the number of neurons in the last layer and select the appropriate activation function. The Image importer accepts images from local files or live stream of a webcam. It can automatically adjust the number of neurons and select the activation function for the last layer. The Custom Tensor importer handles any custom tensor with a predefined format, primarily intended for advanced users.

**Export to Code** Following the creating of a neural network in *asanAI*, the model can be translated into a TensorFlow format for Python integration. This is achieved through an internal process that generates Python code from the TensorFlow.js model. The generated code automates the conversion and execution of the model, and includes annotations on how to apply the model to local data.

The Python script automatically generated through this process dynamically constructs executable code for the TensorFlow model. It also provides guidance on setting up a virtual environment, installing dependencies, and executing the script, thus streamlining the integration into standalone Python projects.

## Implementation Considerations

In order to fulfil the defined goals of this work and address design concepts in the best possible yet technically viable way, certain considerations during the implementation were made. Due to the wide usage of Ternsorflow in the community as well as its low-level access to data structures, to ensure maximum interoperability and provide best compatibility, *asanAI* is using TensorFlow.js library under the hood.

On every change made to the model-related elements of the site, the model is recompiled while retaining the weights (if enabled and possible, i.e., with the same shapes, layer types). These processes are executed in parallel asynchronously wherever possible, so the main interface remains responsive at all times. The execution speed is determined by the computer's hardware and capabilities.

A function is triggered every 200 milliseconds to verify all fields and validate their contents. If a field does not contain a valid value or the number is outside the specified range, the field is marked in red to indicate an error. If a required field is invalid or missing, training is disabled until the field is filled with valid values.

The generated Python codes are in principle strings that are put together depending on the internal data structure of the network. This way, *asanAI* provides a deterministic code generation method that ensures preserving the logic of the exported model.

In the current implementation, for image data, the input shape is automatically set to `[height, width, 3]`, where `height` and `width` are the dimensions of the input image, and `3` represents the three color channels (red, green, and blue). `[nr_of_X_columns]`, where `nr_of_X_columns` is For CSV data, the input shape is set to the number of feature columns in the input data set, and the output shape is the number of output columns in the CSV.

For custom tensor data, the input shape is set to the shape of the input tensor data, excluding the batch size dimension.

For instant language switching, a JSON-based look-up table is used to dynamically change the content of relevant HTML elements without requiring a page reload. This approach is crucial for ensuring a smooth user experience without relying on a persistent internet connection.

## Conclusion and Future Works

*asanAI* presents a significant step towards democratizing education and access to machine learning. By introducing a no-code, browser-based, offline-first toolkit, this work has lowered the barriers to entry for individuals seeking to engage with ML, regardless of their technical expertise or resources. The intuitive interface, offline operation, and seamless interoperability of *asanAI* make it an ideal tool for users ranging from beginners to experts, empowering them to design, train, and test ML models with ease. Furthermore, the educational potential of *asanAI* is significant, as it simplifies the teaching of machine learning concepts in schools by removing all entry barriers, ensuring seamless compatibility with virtually all devices, and functioning without a persistent internet connection. This enables educators to effectively introduce students to the fundamentals of this rapidly evolving field. With its open-source license and commitment to user privacy, *asanAI* represents a significant contribution to the democratization of machine learning, paving the way for a future where the power of AI is accessible to all.

To further enhance *asanAI*, several future developments are planned, including the integration of a hyperparameter optimizer that automatically identifies optimal network architectures and hyperparameters through the Tree Parzen Estimator, a method akin to HyperOpt (Bergstra, Yamins, and Cox 2013). Additionally, a potential improvement involves enabling the export of models to the ONNX (Open Neural Network Exchange) format, which would facilitate their utilization across a diverse array of machine learning frameworks and technologies, extending beyond TensorFlow and TensorFlow.js. Furthermore, to streamline the utilization of the tools developed within *asanAI*, we are in the process of creating a library named *asanai.js*, which will allow users to seamlessly incorporate *asanAI* features directly into any web application.

## License

*asanAI* is released under an MIT-like license, which permits users to freely utilize and modify the tool for their personal projects.

## Acknowledgements

This work was supported by the German Federal Ministry of Education and Research (BMBF, SCADS22B) and the Saxon State Ministry for Science, Culture and Tourism (SMWK) by funding the competence center for Big Data and AI "ScaDS.AI Dresden/Leipzig".

The authors would like to thank Christoph Lehmann, Taras Lazariv, Claudia Adler and many others who supported the development, testing, and optimization of asanAI.

# References

Bergstra, J.; Yamins, D.; and Cox, D. 2013. Making a Science of Model Search: Hyperparameter Optimization in Hundreds of Dimensions for Vision Architectures. In Dasgupta, S.; and McAllester, D., eds., *Proceedings of the 30th International Conference on Machine Learning*, volume 28 of *Proceedings of Machine Learning Research*, 115–123. Atlanta, Georgia, USA: PMLR.

Carney, M.; Webster, B.; Alvarado, I.; Phillips, K.; Howell, N.; Griffith, J.; Jongejan, J.; Pitaru, A.; and Chen, A. 2020. Teachable Machine: Approachable Web-Based Tool for Exploring Machine Learning Classification. In *Extended Abstracts of the 2020 CHI Conference on Human Factors in Computing Systems*, 1–8. ACM.

Lane, D. 2024. Machine Learning for Kids. https://machinelearningforkids.co.uk/. Accessed: 2024-08-15.

LeNail, A. 2019. NN-SVG: Publication-Ready Neural Network Architecture Schematics. *Journal of Open Source Software*, 4(33): 747.

Microsoft. 2021. Lobe - Machine Learning Made Easy. https://www.lobe.ai/. Accessed: 2024-07-02.

Selvaraju, R. R.; Das, A.; Vedantam, R.; Cogswell, M.; Parikh, D.; and Batra, D. 2016. Grad-CAM: Why did you say that? Visual Explanations from Deep Networks via Gradient-based Localization. *CoRR*, abs/1610.02391.

Shiffman, D. 2020. ml5: Friendly Open Source Machine Learning Library for the Web. *Journal of Open Source Software*, 5(45): 2384.

Smilkov, D.; Thorat, N.; Assogba, Y.; Yuan, A.; Kreeger, N.; Yu, P.; Zhang, K.; Cai, S.; Nielsen, E.; Soergel, D.; et al. 2019. TensorFlow.js: Machine Learning for the Web and Beyond. In *Proceedings of the 2nd SysML Conference*.

Zeiler, M. D.; and Fergus, R. 2013. Visualizing and Understanding Convolutional Networks. *CoRR*, abs/1311.2901.